\documentclass[conference]{IEEEtran}
\IEEEoverridecommandlockouts

\usepackage{cite}
\usepackage{amsmath,amssymb,amsfonts}
\usepackage{algorithmic}
\usepackage{graphicx}
\usepackage{textcomp}
\usepackage{xcolor}
\usepackage{amsmath}
\usepackage{multirow}
\usepackage{hyperref}
\newtheorem{theorem}{Theorem}
\def\BibTeX{{\rm B\kern-.05em{\sc i\kern-.025em b}\kern-.08em
    T\kern-.1667em\lower.7ex\hbox{E}\kern-.125emX}}
\begin{document}

\title{VP-NTK: Exploring the Benefits of Visual Prompting in Differentially Private Data Synthesis\\
}

\author{
\IEEEauthorblockN{
Chia-Yi Hsu$^{1}$\qquad Jia-You Chen$^{1}$\qquad Yu-Lin Tsai$^{1}$\qquad Chih-Hsun Lin$^{1}$ \\Pin-Yu Chen$^{2}$\qquad Chia-Mu Yu$^{1}$\qquad Chun-Ying Huang$^{1}$
}
\IEEEauthorblockA{
\textit{$^1$National Yang Ming Chiao Tung University\qquad $^2$IBM Research}}
}

\maketitle

\begin{abstract}
Differentially private (DP) synthetic data has become the \textit{de facto} standard for releasing sensitive data. However, many DP generative models suffer from the low utility of synthetic data, especially for high-resolution images. On the other hand, one of the emerging techniques in parameter efficient fine-tuning (PEFT) is visual prompting (VP), which allows well-trained existing models to be reused for the purpose of adapting to subsequent downstream tasks. In this work, we explore such a phenomenon in constructing captivating generative models with DP constraints. We show that VP in conjunction with DP-NTK, a DP generator that exploits the power of the neural tangent kernel (NTK) in training DP generative models, achieves a significant performance boost, particularly for high-resolution image datasets, with accuracy improving from 0.644$\pm$0.044 to 0.769. Lastly, we perform ablation studies on the effect of different parameters that influence the overall performance of VP-NTK. Our work demonstrates a promising step forward in improving the utility of DP synthetic data, particularly for high-resolution images.
\end{abstract}


\section{Introduction}
Originating in the field of deep learning for natural language processing, prompt engineering has gained popularity as an innovative technique for efficiently using and adapting pre-trained language models for various downstream tasks \cite{liu2023pre}. On the other hand, the original concept of prompt engineering has blossomed and extended to other domains and data types, such as image and computer vision. In particular, visual prompting has been introduced in \cite{bahng2022exploring}, where the method outperforms linear probing (i.e., attaching a trainable linear head to a pre-trained model) when used alongside large-scale vision models, demonstrating its effectiveness. Interestingly, model reprogramming (MR) can be seen as a generalized version of visual prompting (VP), where the well-trained models are reused in a sampling-efficient manner. Specifically, MR involves inserting an input transformation layer and an output mapping layer into a frozen, pre-trained model for fine-tuning downstream tasks. VP in \cite{bahng2022exploring} corresponds to MR when the input transformation consists of a trainable input perturbation and the output mapping corresponds to specified source-target label associations for label inference.

Visual prompting (VP) has been extensively studied in various applications, including image classification \cite{bahng2022exploring}, cross-domain adaptation \cite{neekhara2022cross, tsai2020transfer}, and so on. In this paper, we explore an additional advantageous facet of VP when paired with pre-trained models - its infusion with differential privacy (DP).

In particular, scaling the parameters of generative models in deep learning often leads to better performance on general tasks. However, under the goal of privacy-preserving machine learning, a strict privacy budget can lead to massive noise injection in the training process, rendering the DP generative model useless. Moreover, the demand for high-resolution images often requires that generative models be equipped with high capacity, creating a tension between the demands of privacy, accuracy, and model complexity. Thus, to address these challenges, we aim to answer the following question:
\begin{center}
    \textit{Will incorporating VP with a well-trained generative model offer better privacy-accuracy tradeoff in DP data synthesis?}
\end{center}

In this paper, we provide an affirmative answer, which is empirically supported and comprehensively compared with other state-of-the-art methods. We focus on existing DP data synthesis algorithms since the improvement from applying VP could be directly observable. Our approach includes the application of VP to one of the state-of-the-art DP data synthesis methods, DP-NTK \cite{dp-ntk}. In particular, when VP is applied to DP-NTK, we observe a significant performance improvement over the original method in the downstream classification task of high-resolution images under an identical privacy budget. As a result, our results reveal the advantages of VP in DP data synthesis and offer new use cases and insights for prompt engineering. We summarize our findings as follows: 
\begin{itemize}
    \item We are the first to explore the benefits of VP in DP data synthesis.
    \item We improve the challenge of DP data synthesis in the high-resolution image domain.
    \item We provide insights into how VP could reuse the well-trained generative model. 
\end{itemize}
\vspace{5mm}

\section{Related Works and Background}\label{sec:background}
\textbf{Visual Prompting (VP).
} 
The intent of VP is to fully reuse a pre-trained model to perform a new task, leveraging the pre-trained model's weights without further modification or additional fine-tuning.

VP through a trainable input perturbation is revisited in \cite{bahng2022exploring}, and the authors showed competitive results on a subset of 12 image classification tasks over linear probing and full fine-tuning on pre-trained image classifiers and the CLIP model. We note that in this paper we focus exclusively on VP in the input prompt engineering setting, and leave the alternative setting of layer-wise visual prompt tuning as future work.

\textbf{Differentially Private Data Synthesis.} Since both DPSGD and PATE are common techniques to enforce DP constraints on deep learning models, there are many algorithms for DP generative models that surround both concepts. However, since the implementation of DPSGD requires gradient clipping and noise addition at each training step, this in turn leads to severe information loss in model updates. Therefore, several works are dedicated to reducing the loss, such as \cite{McMahan2018AGA, Thakkar2019DifferentiallyPL, Yu2021DoNL}. On the other hand, there are several works based on PATE such as \cite{ g-pate} which spends the privacy budget on selecting more useful iterations so that the discriminators can learn in the right direction.



Other methods such as DP-MERF \cite{harder2021dpmerf}, DP-NTK \cite{dp-ntk} use various pre-trained perceptual features to further facilitate the DP training process. They achieve this by matching the perturbed data distribution with the generator. Following frameworks similar to DP-MERF, DP-Sinkhorn \cite{Cao2021DPSinkhorn} and PEARL \cite{pearl} train the generator by minimizing Sinkhorn divergence with semi-debiased Sinkhorn loss and the feature distance, respectively. On the other hand, works such as ~\cite{dockhorn2022differentially, wang2024dp} focus on infusing DP with diffusion models, which are state-of-the-art generative models.

\textbf{Visual Prompting with DP.} 
Some recent works that combine VP and DP include Reprogrammable-FL \cite{arif2023reprogrammable} and Prom-PATE \cite{li2023exploring}, where both exploit VP to improve the performance of DP classifiers.

\textbf{High-Resolution Data Synthesis.} 
Ensuring privacy in high-resolution data synthesis is particularly challenging due to information loss caused by DP techniques. Adapting pre-trained models for high-resolution tasks with VP may enhance performance. However, current work focuses only on exploring the benefits of VP in the construction of classifiers~\cite{chen2023understanding}, leaving generative models unexplored.
\section{Main Approach}\label{sec:main}
In this section, we will present the main approach, VP-NTK, as a hybrid of DP generative model and visual prompting. First, we will provide a brief overview of DP-NTK, as this is our choice of DP generative model. Second, we will present how VP can be infused into DP-NTK to create our main approach, VP-NTK, with enhanced capabilities.

\subsection{DP-NTK}
Given a labeled dataset $\{(x_{i}, y_{i})\}^{m}_{i=1} \sim P$ where $m$ represents the number of data and $P$ represents the true data distribution, DP-NTK first construct the true mean embedding of the data distribution as $\hat{\mu}_{P} = \frac{1}{m}\sum_{i=1}^{m} \phi(x_{i})y_{i}^{T}$ where label $y_{i}$ is in the form of one-hot encoding and $\phi(\cdot)$ stands for feature embedding generated from the neural tangent kernel (i.e., $\phi(x) = \frac{\nabla_{\theta} f(x;\theta)}{|| \nabla_{\theta} f(x;\theta) ||}$ where $f(\cdot;\theta)$ and $\theta$ denote the deep network and its corresponding parameter).

Then, to preserve privacy, DP-NTK uses the Gaussian mechanism \cite{dwork2014algorithmic} to obtain a perturbed mean embedding $\tilde{\mu}_{P}$ by 
$
    \tilde{\mu}_{P} = \hat{\mu}_{P} + \mathcal{N}(0, \frac{4\sigma^{2}}{m^{2}}I)
$
where the variance parameter $\sigma$ is controlled by the privacy parameters $\epsilon$ and $\delta$. 

On the other hand, the generator $G$ generates $n$ synthetic data samples $x^{\prime}_{i}$ with standard Gaussian noise $z_{i}$ and generated label $y^{\prime}_{i}$ (i.e. $x^{\prime}_{i} = G(z_{i}, y^{\prime}_{i})$). Similarly, DP-NTK constructs the mean embedding $\hat{\mu}_{Q}$ of the synthetic data as
$
    \hat{\mu}_{Q} = \frac{1}{n}\sum_{i=1}^{n} \phi(x^{\prime}_{i}){y^{\prime}_{i}}^{T}
$

Finally, DP-NTK simulates the true data distribution by minimizing the empirical maximum mean discrepancy between the two embeddings as follows
$
    \widetilde{\text{MMD}}(P,Q) = || \tilde{\mu}_{P} - \hat{\mu}_{Q}|_{F}^{2}
$,
 where $||\cdot||_{F}$ is the Frobenius norm.

\subsection{VP-NTK}
\begin{figure}[hbtp]
    \centering
    \includegraphics[width=0.5\textwidth]{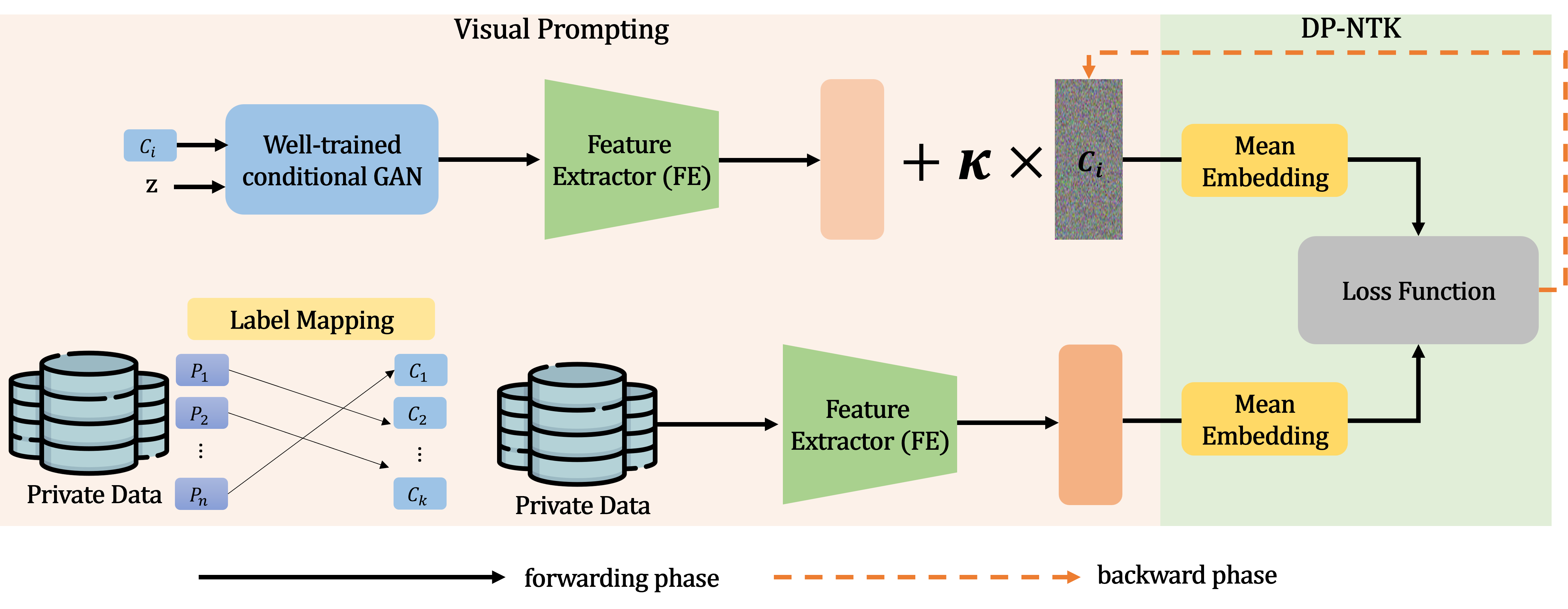}
    \caption{The framework of VP-NTK.}
    \vspace{-4mm}
    \label{fig:framework}
\end{figure}

In our framework, we reuse a well-trained conditional generator and feature extractor as our source models. First, we perform a pre-determined label mapping to determine how classes in the private data should correspond to classes in the conditional generator.
Then, we acquire the features of both the private data and the images generated by the generator by feeding them through a pre-trained feature extractor that remains fixed throughout the training process. Finally, we incorporate VP by adding trainable noise to the features of $G(z, y)$ to learn the distribution of the features of the private data. Note that each class has its own trainable noise, i.e., its own visual stimulus. The role of trainable noise is to better align the pre-trained features to those of the private ones, allowing VP to re-use the private feature space when synthesizing future samples. Finally, we connect the visual prompt feature to the original pipeline of DP-NTK to ensure that the learning process satisfies the DP guarantee. Fig~\ref{fig:framework} illustrates the workflow of VP-NTK.
\begin{table*}[hbtp]
    \caption{Classification accuracy results under $(1, 10^{-5})$-DP  with image resolution as $64\times64$.}
    \centering
    \small
    \addtolength{\tabcolsep}{-2pt}    
    \begin{tabular}{@{}|c|l|ccccccc||c|@{}}
    \hline & \hspace{2ex}$\varepsilon$    & \shortstack{DP-MERF} & \shortstack{G-PATE} & \shortstack{DP-Sinkhorn} & \shortstack{DataLens} & \shortstack{DP-HP} & \shortstack{NDPDC} & \shortstack{PEARL} &\shortstack{VP-NTK } \\ 
    \hline\hline
    {\shortstack{CelebA-Gender}} & $\varepsilon = 1$  & 0.594 & 0.670  & 0.543 & 0.700  & 0.656 & 0.540 & 0.634 & \textbf{0.707}    \\
    & $\varepsilon = 10$      & 0.608    & 0.690 & 0.621 & 0.729 & 0.617 & 0.600 & 0.646 & \textbf{0.769} \\ 

    \hline
    \multirow{2}{*}{\shortstack{CelebA-Hair}}  & $\varepsilon = 1$   & 0.441  & 0.499 & $\times$ & 0.606  & 0.561 & 0.498 & 0.606 & \textbf{0.653} \\
    & $\varepsilon = 10$  & 0.449  & 0.622 & $\times$ & 0.622  & 0.474 & 0.462 &0.626 & \textbf{0.641} \\ 
    \hline
    \end{tabular}
    \label{tab:classification}
\end{table*}

We explain the details in the workflow. For label mapping, we randomly map classes of private data to the predefined conditional signals of the generator (e.g., "dog" in the source to "man" in the private data). The use of label mapping in VP-NTK allows private data to have labels mapped to those of the pre-trained data, making VP available to reuse large pre-trained generators to synthesize better data. Furthermore, since the label mapping is determined randomly before any training process, this ensures that VP-NTK could remain DP, similar to that of DP-NTK.

When adding noise to the features, the coefficient $\kappa$ is used to control the amount of noise. On the other hand, the original loss function of DP-NTK is to compute the empirical maximum mean discrepancy between true and synthetic data. Instead of relying solely on Maximum Mean Discrepancy (MMD), we also incorporate cosine similarity into our loss. Finally, we introduce the coefficient $\alpha$, which controls a penalty term on the norm of the trainable input noise to avoid over-fitting. Note that the source model remains frozen in our training process, including the well-trained conditional generator and the feature extractor itself.

\begin{theorem}
    VP-NTK satisifies $(\epsilon, \delta)$-DP.
\end{theorem}
\textit{Proof}: Due to limited space, we defer the proof to web for demo. \footnote{https://anonymous.4open.science/r/ICASSP2025-FBFE/Theorem1.png}

\section{Experiments}\label{sec:experiment}
In this section, we conduct experiments on complicated datasets to demonstrate the performance of our proposed VP-NTK. Furthermore, we present the ablation study to show the rationale behind the hyperparameter selection.

\subsection{Experiment Setup}
We demonstrate the utility of VP-NTK on CelebA datasets at different resolutions, including $64\times64$, $128\times128$, and the high-resolution $256\times256$. Derived from CelebA, we generated two additional datasets: CelebA-Gender for binary gender classification and CelebA-Hair for multiclass hair color classification (including black, blonde, and brown). Both are colorful, increasing the challenge of preserving the utility of the synthetic data. For the DP guarantee, we set $(\epsilon, \delta)$ to $(1, 10^{-5})$ and $(10, 10^{-5})$. All experiments are done on NVIDIA GeForce RTX 4090. In the following, we present the baseline, the VP-NTK settings, and how to evaluate the synthetic data. 
\begin{itemize}
    \item \textbf{Baselines.}\;\; We compare VP-NTK with state-of-the-art DP image synthesis methods such as DP-MERF \cite{harder2021dpmerf}, DataLens \cite{wang2021datalens}, G-PATE \cite{g-pate}, DP-Sinkhorn \cite{Cao2021DPSinkhorn}, DP-HP \cite{dp-hp}, Nonlinear DPDC (NDPDC) \cite{dpdc}, and PEARL \cite{pearl}. All comparisons are based on their official codes. The choice of baseline methods considers some classical and novel methods in DP generation. For instance, G-PATE transforms the original PATE scheme of DP classification into its generative counterparts, while NDPDC utilizes dataset condensation to generate DP synthetic samples.
    \item \textbf{VP-NTK.}\;\; Our framework includes a pre-trained GAN and a feature extractor (FE). For all experiments, we use the IC-GAN \cite{casanova2021instanceconditioned} and employ the ResNet18 as the FE, pre-trained on the Tiny-ImageNet dataset. There are four important hyper-parameters in our experiment which are $\kappa$, $\eta$, $\alpha$, and the choice of loss functions. $\eta$ represents the learning rate. For the standard solution of our experiments, we set $\kappa=16$, $\eta=10^{-2}$, $\alpha=0.05$, and the loss function combines MMD and cosine similarity. 
    \item \textbf{Evaluation.}\;\; All synthetic data generated by each method are used to train classifiers with the same architecture. We compare the test accuracy on the test sets of CelebA-Gender and CelebA-Hair, where high test accuracy represents higher utility.
\end{itemize}



\subsection{Comparison with Existing DP Generative Models}

We compare VP-NTK with several SOTA DP generative models. Table~\ref{tab:classification} shows the result of our method and other reference methods on 64$\times$64 image resolution. Here we note that while we also experiment on images with higher resolutions, such as 128$\times$128 and 256$\times$256, existing methods present no data for further comparison. As can be seen from Table~\ref{tab:classification}, VP-NTK has outstanding performance against all GAN-based DP generative models and an accuracy gap of more than 4$\%$ with $\epsilon=10$ on CelebA-Gender.
\par For other larger image resolutions, such as 128$\times$128 and 256$\times$256, Table~\ref{tab:result} shows that our method also performs well on them. However, for higher resolution data, other methods require DP image generators with high capacity(i.e., large parameter counts), which will result in worsening performance than the result in Table~\ref{tab:classification}, while VP-NTK benefits from the potential of visual prompting, allowing the DP generative model to handle data with increasing resolution.

\begin{table}[hbtp]
    \caption{Accuracy for different dataset and $\epsilon$}
\vspace{-3mm}
    \begin{center}
    \resizebox{\columnwidth}{!}{
    \begin{tabular}{|c|c|c|c|} 
     \hline
     \textbf{Dataset}       & \textbf{Image size} & \textbf{$\varepsilon$} & \textbf{Accuracy$\pm$Std(\%)} \\
     \hline \hline
     \textbf{CelebA-Gender} & 128$\times$128 & 1  & 79.24$\pm$0.38 \\
     \textbf{ }             & 256$\times$256 &    & 82.67$\pm$1.75 \\
     \hline
     \textbf{CelebA-Gender} & 128$\times$128 & 10 & 79.01$\pm$0.41 \\
     \textbf{ }             & 256$\times$256 &    & 82.28$\pm$1.98 \\
     \hline
     \textbf{CelebA-Hair}   & 128$\times$128 & 1  & 64.14$\pm$0.17 \\
     \textbf{ }             & 256$\times$256 &    & 65.48$\pm$0.30 \\
     \hline
     \textbf{CelebA-Hair}   & 128$\times$128 & 10 & 63.66$\pm$0.41 \\
     \textbf{ }             & 256$\times$256 &    & 65.85$\pm$0.35 \\
     \hline
    \end{tabular}
    }
    \end{center}
    \vspace{-5mm}
    \label{tab:result}
\end{table}

\subsection{Ablation Study}
For all the experiments in this section, we evaluated the performance on the CelebA-Gender and CelebA-Hair datasets with a resolution of 128$\times$128 and a privacy budget of $\varepsilon = 1$. \\
\par \textbf{The Effect of $\kappa$.} The $\kappa$ is used to control the amount of noise. The table \ref{tab:kappa(cof)} shows how $\kappa$ affects performance. If the value of $\kappa$ is too low, for example, $\kappa = 2$, it may be difficult to effectively transfer synthetic features to private data features. On the other hand, if $\kappa$ is too large, it can lead to overfitting. There are the same trends between CelebA-Gender and Hair that the performances are better as $\kappa$ in a reasonable range (e.g., $\kappa=4,8,16$) shown in table \ref{tab:kappa(cof)}.

\vspace{-3mm}
\begin{table}[hbtp]
    \caption{Accuracy for different $\kappa$}
    \begin{center}
    \resizebox{0.75\columnwidth}{!}
    {
    \begin{tabular}{|c|c|c|} 
         \hline
         \textbf{Dataset}         & \textbf{$\kappa$} & \textbf{Accuracy$\pm$Std(\%)} \\
         \hline \hline
         \textbf{}                  & 2  & 77.31$\pm$0.15 \\
         \textbf{}                  & 4  & 79.23$\pm$0.31 \\
         \textbf{CelebA-Gender}     & 8  & 79.28$\pm$0.54 \\
         \textbf{}                  & 16 & 79.24$\pm$0.38 \\
         \textbf{}                  & 32 & 78.70$\pm$0.55 \\
         \hline
         \textbf{}                  & 2  & 53.43$\pm$0.06 \\
         \textbf{}                  & 4  & 60.47$\pm$0.40 \\
         \textbf{CelebA-Hair}       & 8  & 63.05$\pm$0.48 \\
         \textbf{}                  & 16 & 63.87$\pm$0.05 \\
         \textbf{}                  & 32 & 64.12$\pm$1.06 \\
         \hline
    \end{tabular}
    }
    \end{center}
    \label{tab:kappa(cof)}
\end{table}

\par \textbf{The Effect of $\eta$.} Table~\ref{tab:eta} shows the result on CelebA-Gender 128$\times$128. Similarly, Table~\ref{tab:eta} shows the result on CelebA-Hair 128$\times$128. Obviously, for both results, using $\eta = 10^{-2}$ has the best performance. This is because scaling $\eta$ leads to an unstable result, while decreasing the learning rate leads to slower convergence and thus harder search for the optimal visual prompt.

\begin{table}[h!]
    \caption{Accuracy for different $\eta$}
\vspace{-2mm}
    \begin{center}
    \resizebox{0.75\columnwidth}{!}
    {
    \begin{tabular}{|c|c|c|} 
         \hline
         \textbf{Dataset}        & \textbf{$\eta$} & \textbf{Accuracy$\pm$Std(\%)} \\
         \hline \hline
         \textbf{}               & 1e-5 & 52.81$\pm$2.26 \\
         \textbf{}               & 1e-4 & 46.77$\pm$3.26 \\
         \textbf{CelebA-Gender}  & 1e-3 & 79.22$\pm$0.75 \\
         \textbf{}               & 1e-2 & \textbf{79.24$\pm$0.38} \\
         \textbf{}               & 0.1  & 79.19$\pm$0.86 \\
         \textbf{}               & 1    & 77.03$\pm$3.02 \\
         \hline
         \textbf{}               & 1e-5 & 29.33$\pm$0.79 \\
         \textbf{}               & 1e-4 & 33.31$\pm$1.56 \\
         \textbf{CelebA-Hair}    & 1e-3 & 63.60$\pm$0.27 \\
         \textbf{}               & 1e-2 & \textbf{63.87$\pm$0.05} \\
         \textbf{}               & 0.1  & 58.83$\pm$4.42 \\
         \textbf{}               & 1    & 58.16$\pm$3.32 \\
         \hline
    \end{tabular}
    }
    \end{center}
    \vspace{-3mm}
    \label{tab:eta}
\end{table}

\textbf{The Effect of $\alpha$.} We evaluate the effect of $\alpha$ on both datasets, with the results presented in Table~\ref{tab:alpha}. As can be seen, $\alpha$ does not have a large effect on the overall result. This is mainly due to the small magnitude of the visual cues compared to either cosine similarity or MMD loss.

\textbf{The Effect of Different Loss.}
We evaluate the effect of different losses with the results shown in Table~\ref{tab:loss}. Although MMD could provide theoretical guarantees between the true and synthetic distribution. However, the empirical MMD used in both VP-NTK and DP-NTK \cite{dp-ntk} could sometimes differ from the true MMD. Therefore, we include cosine similarity as an alternative metric and discovered that mixing in equal proportions will offer the best result.

\begin{table}[h!]
    \caption{Accuracy for different $\alpha$}
\vspace{-3mm}
    \begin{center}
    \resizebox{0.75\columnwidth}{!}
    {
    \begin{tabular}{|c|c|c|} 
         \hline
         \textbf{Dataset}        & \textbf{$\alpha$} & \textbf{Accuracy$\pm$Std(\%)} \\
         \hline \hline
         \textbf{}               & 0.01 & 78.31$\pm$0.63 \\
         \textbf{CelebA-Gender}  & 0.05 & 79.24$\pm$0.38 \\
         \textbf{}               & 0.1  & 79.44$\pm$0.37 \\
         \textbf{}               & 1    & 79.25$\pm$0.16 \\
         \hline
         \textbf{}               & 0.01 & 63.69$\pm$1.72 \\
         \textbf{CelebA-Hair}    & 0.05 & \textbf{63.87$\pm$0.05} \\
         \textbf{}               & 0.1  & 63.41$\pm$0.46 \\
         \textbf{}               & 1    & 57.57$\pm$2.62 \\
         \hline
    \end{tabular}
    }
    \end{center}
    \vspace{-5mm}
    \label{tab:alpha}
\end{table}

\begin{table}[h!]
    \caption{Accuracy for different loss}
\vspace{-3mm}
    \begin{center}
    \resizebox{0.7\columnwidth}{!}
    {
    \begin{tabular}{|c|c|c|} 
         \hline
         \textbf{Dataset}        & \textbf{loss} & \textbf{Accuracy$\pm$Std(\%)} \\
         \hline \hline
         \textbf{}  & MMD    & 58.08$\pm$0.27 \\
         \textbf{CelebA-Hair}               & mixed  & \textbf{79.24$\pm$0.38} \\
         \textbf{}               & cosine & 79.21$\pm$0.35 \\
         \hline
         \textbf{}    & MMD    & 36.65$\pm$2.12 \\
         \textbf{CelebA-Hair}              & mixed  & \textbf{63.87$\pm$0.05} \\
         \textbf{}               & cosine & 63.83$\pm$0.07 \\
         \hline
    \end{tabular}
    }
    \end{center}
    \vspace{-5mm}
    \label{tab:loss}
    \end{table}

\label{sec:experiment}

\section{Conclusion}\label{sec:conclusion}
In this paper, we've explored the preliminary benefits of integrating visual prompting into the DP data synthesis pipeline. Specifically, when applied to one of the SOTA DP generative models, we can observe an increase in accuracies when the generative model is used for a downstream classification task. Furthermore, the integrated DP generative model could also address the challenge of generating high-resolution images. Our discovery revealed that VP is a promising method to accelerate further research in constructing DP generative models that improve the privacy-utility tradeoff.

\bibliographystyle{IEEEbib}
\bibliography{refs}


\end{document}